\documentclass{article}
\usepackage{ijcai21}

\usepackage{times}
\usepackage{soul}
\usepackage{url}
\usepackage[hidelinks]{hyperref}
\usepackage[utf8]{inputenc}
\usepackage[small]{caption}
\usepackage{graphicx}
\usepackage{amsmath}
\usepackage{amsthm}
\usepackage{booktabs}
\usepackage{appendix}
\usepackage{amssymb}
\usepackage{algorithm}% http://ctan.org/pkg/algorithms
\usepackage{algpseudocode}% http://ctan.org/pkg/algorithmicx
\algnewcommand{\LineComment}[1]{\Statex \(\triangleright\) #1}
\usepackage{multirow}
\usepackage{makecell}
\usepackage{bbding}
\usepackage{array}
\usepackage{etoolbox}

\BeforeBeginEnvironment{appendices}{\clearpage}
\newtheorem{corollary}{Corollary} 
\newtheorem{lemma}{Lemma}

\newcolumntype{L}[1]{>{\raggedright\arraybackslash}p{#1}}
\newcolumntype{R}[1]{>{\raggedleft\arraybackslash}p{#1}}

\title{Learning Unknown from Correlations: Graph Neural Network for Inter-novel-protein Interaction Prediction}

\author{
Guofeng Lv
\and
Zhiqiang Hu \and
Yanguang Bi \and
Shaoting Zhang
\affiliations
SenseTime Research\\
\emails
\{lvguofeng, huzhiqiang, biyanguang, zhangshaoting\}@sensetime.com
}

\begin{document}

\maketitle

\begin{abstract}
  The study of multi-type Protein-Protein Interaction (PPI) is fundamental for understanding biological processes from a systematic perspective and revealing disease mechanisms. Existing methods suffer from significant performance degradation when tested in unseen dataset. In this paper, we investigate the problem and find that it is mainly attributed to the poor performance for inter-novel-protein interaction prediction. However, current evaluations overlook the inter-novel-protein interactions, and thus fail to give an instructive assessment. As a result, we propose to address the problem from both the evaluation and the methodology. Firstly, we design a new evaluation framework that fully respects the inter-novel-protein interactions and gives consistent assessment across datasets. Secondly, we argue that correlations between proteins must provide useful information for analysis of novel proteins, and based on this, we propose a graph neural network based method (GNN-PPI) for better inter-novel-protein interaction prediction. Experimental results on real-world datasets of different scales demonstrate that GNN-PPI significantly outperforms state-of-the-art PPI prediction methods, especially for the inter-novel-protein interaction prediction.\footnote{Codes are available at https://github.com/lvguofeng/GNN\_PPI.}
\end{abstract}

\section{Introduction}
\label{section:1}

Protein-protein Interactions (PPIs) play an important role in most biological processes. In addition to direct physical binding, PPI also has many other, indirect ways of co-operation and mutual regulation, such as exchange reaction products, participate in signal relay mechanisms, or jointly contribute toward specific organismal functions \cite{szklarczyk2016string}. It can be said that the study of PPIs and their interaction types are essential toward understanding cellular biological processes in normal and disease states, which in turn facilitate the therapeutic target identification and novel drug design \cite{skrabanek2008computational}. There are many experimental methods to detect PPI, where the most conventional and widely used high-throughput methods are yeast two-hybrid screening \cite{fields1989novel}. However, the experiment-based methods are expensive and time-consuming, but more importantly, even if a single experiment has detected PPI, it cannot fully interpret its types \cite{de2010protein}. Evidently, we urgently need reliable computational methods that are learned from the accumulated PPI data to predict the unknown PPIs accurately.

\begin{figure}[tb]
\centering
\includegraphics[width=7cm]{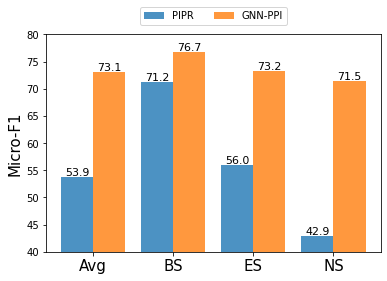}
\caption{Results of PIPR (baseline) and GNN-PPI (ours) when trained on the smaller dataset SHS148k and tested on the larger STRING dataset. The metric is micro F1 score for multi-label PPI type prediction. Avg is the overall result of the testset. For further investigation, we divide the testset into BS, ES and NS subsets, where BS denotes Both of the pair proteins in interaction were Seen during training, ES denotes Either (but not both) of the pair proteins was Seen, and NS denotes Neither proteins were Seen during training. We regard ES and NS as inter-novel-protein interactions.}
\label{fig:picture000}
\end{figure}

Despite long-term research works \cite{guo2008using,hashemifar2018predicting,chen2019multifaceted} make noticeable progress, existing methods suffer from significant performance degradation when tested on unseen dataset. Take the state-of-the-art model PIPR \cite{chen2019multifaceted} as an example, compared tested on trainset-homologous SHS148k testset with on a larger STRING testset, micro F1 score drops from 92.42 to 53.85. For further investigation, we divide the STRING testset into BS, ES and NS subsets, where BS denotes Both of the pair proteins in interaction were Seen during training, ES denotes Either (but not both) of the pair proteins was Seen, and NS denotes Neither proteins were Seen during training. As clearly shown in Figure \ref{fig:picture000}, poor performance in the ES and NS subsets (collectively termed as inter-novel-protein interactions in this paper) is the main reason for the performance degradation.

On the other hand, current evaluations on the trainset-homologous SHS148k testset apply a protein-irrepective per-interaction randomized strategy to divide the trainset and testset, and consequently, BS comprises over 92\% of the whole testset and dominates the overall performance (see Appendix A for more discussions). The evaluations overlook the inter-novel-protein interactions, and are thus not instructive for the performance when tested on other datasets. As a result, in this paper we firstly design a new evaluation framework with two per-protein randomized data partition strategies. Instead of simple protein-independent randomization, we take also into consideration the distance between proteins and utilize Breadth-First and Depth-First Search methods to construct the testset. Comparison experiments between the trainset-homologous testset and the unseen STRING testset demonstrates the proposed evaluation can give consistent assessment across datasets.

Besides the evaluation, for the methodology existing works take PPIs as independent instances. Correlations between proteins have long been ignored.  Intuitively, for predicting the type of interaction between protein A and B, the interaction between protein A and C, as well as B and C must provide useful information. The correlations can be naturally modeled and excavated with a graph, where proteins serve as the nodes and interactions as the edges. In this paper, the graph is processed with a graph neural network based model (GNN-PPI). As demonstrated in Figure \ref{fig:picture000}, the introduction of correlations and the proposed GNN-PPI model have largely narrow the performance gap between BS, ES and NS subsets.

In summary, the contribution of this paper is three-fold:
\begin{enumerate}
    \item We design a new evaluation framework that fully respects the inter-novel-protein interactions and give consistent assessment across datasets.
    \item We propose to incorporate correlation between proteins into the PPI prediction problem. A graph neural network based method is presented to model the correlations.
    \item The proposed GNN-PPI model achieves state-of-the-art performance in real datasets of different scales, especially for the inter-novel-protein interaction prediction.
\end{enumerate}

\section{Related Work}

The primary amino acid sequences are confirmed to contain all the protein information \cite{anfinsen1972formation} and are extremely easy to obtain. Thus, there is a longstanding interest in using sequence-based methods to model protein-related tasks. The research work of PPI prediction and classification can be summarized into two stages. The early research is based on Machine Learning (ML) \cite{guo2008using,wong2015detection,silberberg2014method,shen2007predicting}. These methods provide feasible solutions, but their performance is limited by the PPI feature representation and model expressiveness. Deep Learning (DL) has recently been widely used in bioinformatics problems due to its powerful expressive ability, including PPI prediction and classification. These works \cite{li2018deep,hashemifar2018predicting,chen2019multifaceted,sun2017sequence} typically use Convolution Neural Networks or Recurrent Neural Networks to extract features from the amino acid sequence of the protein.

% Recently, many sequence-based computational methods have been proposed, which provide critical solutions for the PPI prediction and classification tasks. Early works address the tasks with machine learning models, such as support vector machine(SVM) \cite{guo2008using,shen2007predicting}, Random Forest(RF) \cite{wong2015detection}, and logistic regression(LR) \cite{silberberg2014method}. These methods simply concatenate the features of the two proteins as PPI features. These features include CT \cite{shen2007predicting}, AC \cite{guo2008using}, and CTD \cite{du2017deepppi}. 

% Compared with classic machine learning methods, deep learning methods are advantaged in extracting features directly from data and capture nonlinear dependencies between abstract features. Therefore, deep learning methods have been unprecedentedly popular in recent years and have been successfully applied to various problems in bioinformatics \cite{min2017deep}. \cite{li2018deep} uses two separated CNN to encode protein features and then concatenate them to generate PPI features. On the contrary, \cite{hashemifar2018predicting} proposes a Siamese CNN architecture and a novel random projection module to extract the features of protein pairs. \cite{chen2019multifaceted} proposes a framework based on CNN and RNN for PPI prediction and classification. 

More recent work has focused on the feature representation of proteins. \cite{saha2020amalgamation} proposes a novel deep multi-modal architecture, which extracts multi-modal information from protein structure and existing text information in biomedical literature. \cite{nambiar2020transforming} proposes a Transformer based neural network to generate proteins pre-trained embedding. In the latest research, \cite{yang2020graph} considers the correlation of PPIs and first proposed to use GCN\cite{kipf2016semi} to learn protein features in the PPI network automatically. However, their work cannot be extended to the multi-label PPI classification.
% Their experimental results prove that using the pre-train embedding in PPI prediction tasks can effectively improve performance.

To the best of our knowledge, the current work of PPI has not been concerned with the problems of inter-novel-protein interactions. However, In the field of Drug-drug Interaction (DDI), \cite{deng2020multimodal} mentioned that the testset is divided according to whether the drug was seen during training, and the results show that the performance for the inter-novel-drug interactions is extremely degraded, but the original paper does not propose a solution. 

% For correlativity, it has been proven that the structural information of PPI networks is beneficial in protein-related tasks\cite{peng2017protein}. Thanks to the development of graph neural networks in recent years, these methods have become a paradigm for representing learning on graphs. 

\section{Methodology}

% In this section, we first formulate the few-shot learning for multi-label PPI prediction problem (Section 3.1). Then we provide an overview of the proposed GNN-PPI framework (Section 3.2). After that, we introduce our proposed PPI partition strategy based on few-shot learning(Section 3.3). Finally, we divide into two stages to introduce our proposed algorithm GNN-PPI in detail: protein feature coding and multi-label PPI prediction(Section 3.4 and 3.5).
\begin{figure*}[ht]
\centering
\includegraphics[width=15cm]{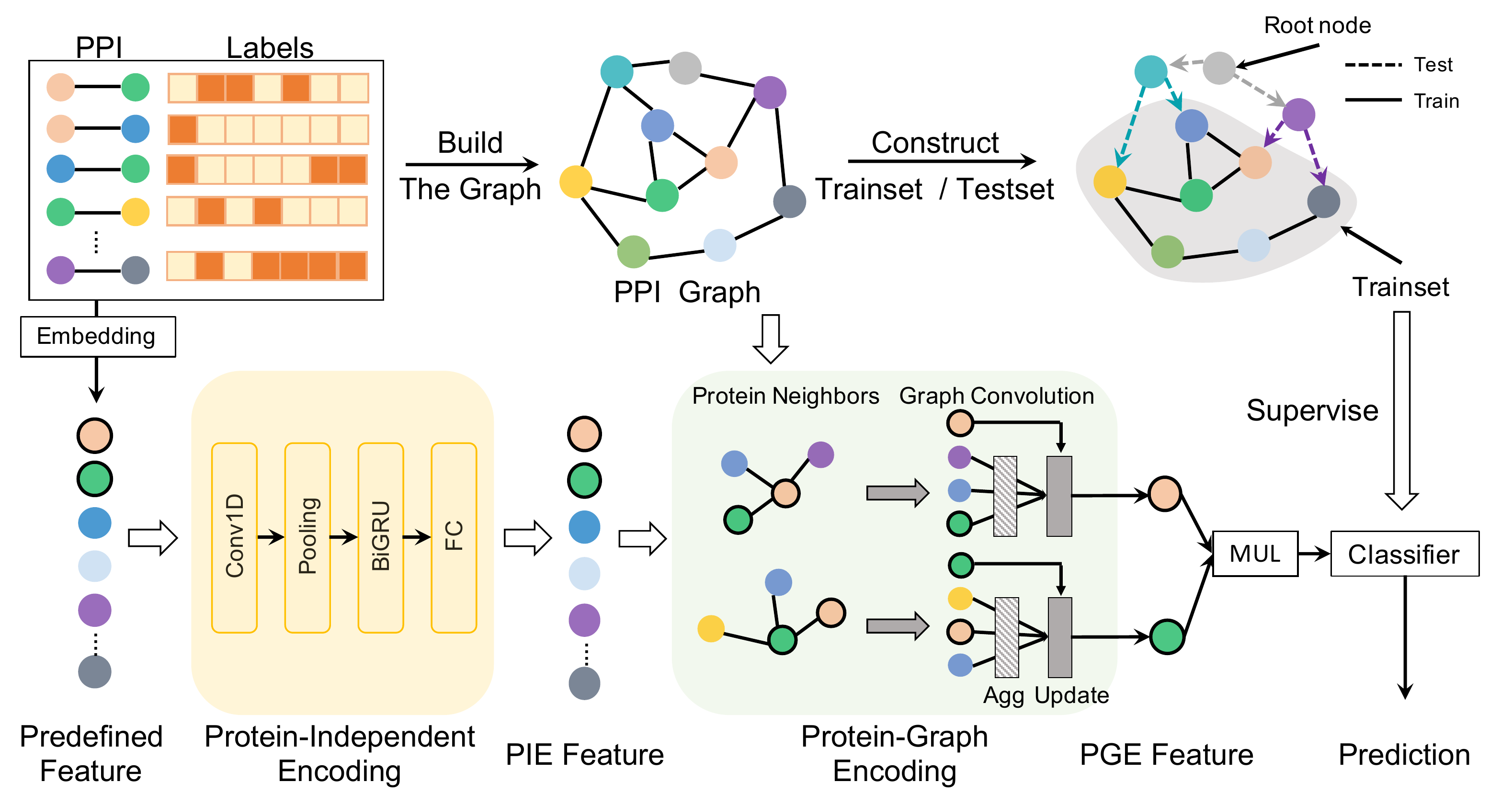}
\caption{Development and evaluation of the GNN-PPI framework. Pairwise interaction data are firstly assembled to build the graph, where proteins serve as the nodes and interactions as the edges. The testset is constructed by firstly selecting the root node and then performing the proposed BFS or DFS strategy. The model is developed by firstly performing embedding for each protein to obtain predefined features, then processed by Convolution, Pooling, BiGRU and FC modules to extract protein-independent encoding (PIE) features, which are finally aggregated by graph convolutions and arrive at protein-graph encoding (PGE) features.  Features of the pair proteins in interaction are multiplied and classified, supervised by the trainset labels.}

\label{fig:picture001}
\end{figure*}

\subsection{Problem Formulation}

% \subsubsection{Multi-Lable PPI.} 
Suppose we have protein set $\mathcal{P}=\{p_0, p_1, ..., p_{n}\}$ and PPI set $\mathcal{X}=\{x_{ij}=\{p_i, p_j\} | i \neq j, p_i, p_j \in \mathcal{P}, I(x_{ij})\in \{0, 1\}\}$. $I$ is the PPI indicator function, if $I(x_{ij}) = 1$, then it means that protein $p_i$ interacts with protein $p_j$. Note that when $I(x_{ij}) = 0$, it may mean that protein $p_i$ and $p_j$ will not interact, or they have a potential interaction while it has not been discovered so far. In order to avoid unnecessary errors, we will not do any operation on unknown protein pairs (default $\forall x_{ij} \in \mathcal{X}, I(x_{ij}) = 1$). We define PPI label space as $\mathcal{L}=\{l_0, l_1, ..., l_n\}$ with $n$ possible interaction types. For each PPI $x_{ij}$, its labels is represented as $y_{ij} \subseteq \mathcal{L}$. In summary, the multi-type PPI dataset is defined as $\mathcal{D} = \{(x_{ij}, y_{ij}) | x_{ij} \in \mathcal{X} \}$. Considering the correlation of PPIs, we use protein as nodes and PPIs as edges to build the PPI graph $\mathcal{G}=(\mathcal{P}, \mathcal{X})$.

% \subsubsection{Few-shot Learning.} 

The task of multi-type PPI learning is to learn a model $\mathcal{F}:x \to \hat{y}$ from the training set $\mathcal{X} _{\mathrm{train}}$. For any protein pair $x_{ij} \in \mathcal{X} _{\mathrm{test}}$, the model $\mathcal{F}$ predict $\hat{y}_{ij}$ as the set of proper labels for $x_{ij}$. The above-mentioned $\mathcal{X} _{\mathrm{train}}$ and $\mathcal{X} _{\mathrm{test}}$ are obtained from $\mathcal{X}$ based on the evaluation, where $\mathcal{X} _{\mathrm{train}}+\mathcal{X} _{\mathrm{test}}=\mathcal{X}$. Further, according to whether protein $p$ was seen during training, the protein set $\mathcal{P}$ is divided into known $\mathcal{P} _v = \bigcup \mathcal{X}_{\mathrm{train}}$ and unknown $\mathcal{P} _u = \mathcal{P} - \mathcal{P} _{\mathrm{train}}$. Moreover, as mentioned in section \ref{section:1}, $\mathcal{X} _{\mathrm{test}}$ can be divided into $\mathcal{X} _{\mathrm{BS}}, \mathcal{X} _{\mathrm{ES}}$, and $\mathcal{X} _{\mathrm{NS}}$, which defined as follows:
\begin{align*}
\mathcal{X} _{\mathrm{BS}} & = \{x_{ij}| x_{ij}\in \mathcal{X}_{\mathrm{test}}, p_i, p_j \in \mathcal{P} _v\} \\
\mathcal{X} _{\mathrm{ES}} & = \{x_{ij}| x_{ij}\in \mathcal{X}_{\mathrm{test}}, p_i \in \mathcal{P} _u , p_j \in \mathcal{P} _v \\ 
 & \qquad \quad \quad \quad \quad \quad \mathrm{or} \  p_j \in \mathcal{P} _u , p_i \in \mathcal{P} _v \}  \\
\mathcal{X} _{\mathrm{NS}} & = \{x_{ij} | x_{ij}\in \mathcal{X}_{\mathrm{test}}, p_i, p_j \in \mathcal{P} _u\}
\end{align*}
% $$\mathcal{X} _{\mathrm{BS}} = \{x_{ij}| x_{ij}\in \mathcal{X}_{\mathrm{test}}, p_i, p_j \in \mathcal{P} _v\}$$
% $$\mathcal{X} _{\mathrm{ES}} = \{x_{ij}| x_{ij}\in \mathcal{X}_{\mathrm{test}}, p_i \in \mathcal{P} _u \; \mathrm{or} \; p_j \in \mathcal{P} _u\}$$ 
% $$\mathcal{X} _{\mathrm{NS}} = \{x_{ij} | x_{ij}\in \mathcal{X}_{\mathrm{test}}, p_i, p_j \in \mathcal{P} _u\}$$
% The Few-shot learning for multi-label PPI prediction problem (FSL-MPPI) is formulated as follows. Given the Protein set $\mathcal{P}$ and PPI set $\mathcal{X}$, we use the data partition strategy $\mathcal{S}$ to divide $\mathcal{X}$ into $\mathcal{X} _{train}$ and $\mathcal{X} _{test}$, where $\mathcal{X} _{test}$ satisfies $|\mathcal{X} _{bs}| \ll |\mathcal{X} _{es}| + |\mathcal{X} _{ns}|$. FSL-MPPI requires the model $\mathcal{F}$ learned from $\mathcal{X} _{train}$ has good scalability and can accurately predict the multi-label label of PPI in $\mathcal{X} _{test}$.
Since inter-novel-protein interactions are the main bottlenecks, we require the testset $\mathcal{X} _{\mathrm{test}}$ of the evaluation framework to meet condition $|\mathcal{X} _{\mathrm{BS}}| \ll |\mathcal{X} _{\mathrm{ES}}| + |\mathcal{X} _{\mathrm{NS}}|$. Our goal is that under this evaluation, the model $\mathcal{F}$ learned from $\mathcal{X} _{\mathrm{train}}$ can accurately predict the multi-label label of PPI in $\mathcal{X} _{\mathrm{test}}$.

\subsection{Overview}

The GNN-PPI framework and evaluation are shown in Figure \ref{fig:picture001}. We will introduce GNN-PPI from the following three aspects. First is \textbf{Evaluation Framework}. We propose two sets of heuristic data partition schemes based on the PPI network, and the generated testset meets the conditions $|\mathcal{X} _{\mathrm{BS}}| \ll |\mathcal{X} _{\mathrm{ES}}| + |\mathcal{X} _{\mathrm{NS}}|$. Secondly, \textbf{Protein feature encoding}. We design Protein-Independent Encoding (PIE) and Protein-Graph Encoding (PGE) modules to encode protein features. The last is \textbf{Multi-label PPI prediction}. For unknown PPIs, we combine their protein feature encoded by the previous process, calculate their scores in different PPI types, and output its multi-label prediction. 
% It is worth noting that the central idea of GNN-PPI is to consider the PPI Network Graph $G$ structure, by using graph neural network(GNN) to encode the protein. Therefore, we design GNN-PPI as a three-step framework for multi-label PPI prediction:

% \begin{enumerate}
%     \item \textbf{Data Partition Strategy}: Dividing $\mathcal{X}$ for training and testing based on PPI network graph.
%     \item \textbf{Protein feature encoding}: Encoding protein features based on GNN.
%     \item \textbf{Multi-label PPI prediction}: Predicting the multi-label PPI based on protein pair feature learned from the previous step.
% \end{enumerate}

\begin{figure}[tb]
\centering
\includegraphics[width=8cm]{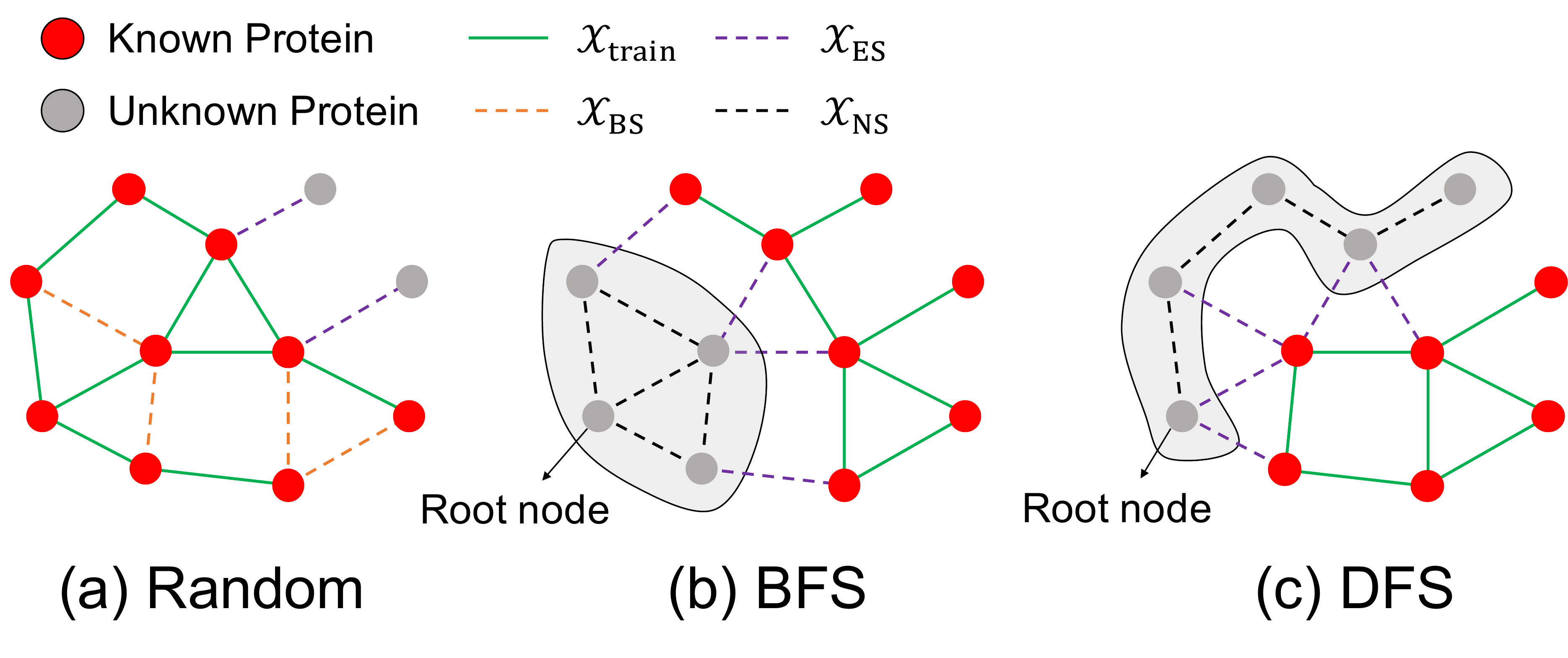}
\caption{Examples of different testset construction strategies. Random is the current scheme, while Breath-First Search (BFS) and Depth-First Search (DFS) are the proposed schemes.}
\label{fig:picture002}
\end{figure}

\subsection{Evaluation Framework}
\label{section:3.3}

Generally, existing machine learning algorithms usually randomly divide part of the dataset as a testset to evaluate the performance of the model. However, in the PPI-related tasks, we have the following Corollary, derived from Erdős–Rényi(ER) random graph model. \cite{erdds1959random,erdHos1960evolution}:
\begin{corollary}
\label{corollary:1}
Randomly divide the PPI dataset, select $t \leq 0.2$ as the testset, then most of the proteins in the testset were seen in training.
\end{corollary}
The detailed proof of the corollary is elaborated in the Appendix A. It can be inferred from this corollary that the performance of the testset obtained by random division only reflects the predictive ability of the PPI between known $\mathcal{P} _v$.
% Generally, in order to evaluate the scalability performance of a model, part of the dataset needs to be divided as a test set. Existing machine learning algorithms usually adopt a random data partition strategy because of the assumption that the training data follows an independent and identical distribution (IID). However, in the multi-label PPI tasks, if the test set is randomly divided(See Figure \ref{fig:picture002}.(c)), there is no unknown PPI in the test set. After we count the proportions of different test sets after random division, on average, the proportion of $\mathcal{X} _{test1}$ will always reach more that 92\%(shown in Table \ref{tab:3}). 
In the real world, there are still many proteins and their PPIs that have not been discovered. We perform empirical studies by comparison of two different time points, 2021/01/25 and 2020/04/11, of the Homo sapiens subset of BioGRID\footnote{https://thebiogrid.org/}\cite{stark2006biogrid} database. We found that the newly discovered proteins exhibit some BFS-like or DFS-like local patterns. Even if the PPIs have been discovered, most of their types remain relatively unexplored. Therefore, we need a brand-new evaluation that can reflect the model's predictive performance on the inter-novel-protein interactions. The next content will introduce the evaluation framework we design. 

We design two heuristic evaluation schemes based on the PPI network, namely \textbf{BFS} and \textbf{DFS}. They simulated two scenarios of unknown proteins $\mathcal{P}_u$ in reality: 

\begin{enumerate}
    \item $\mathcal{P}_u$ interact tightly with each other, and they exist in the form of clusters in the PPI network (See Figure \ref{fig:picture002} (b)).
    \item $\mathcal{P}_u$ are sparsely distributed in the PPI network and have little interaction with each other (See Figure \ref{fig:picture002} (c)).
\end{enumerate}
% Data Partition Strategy we propose is dependent on the PPI network. The generated $\mathcal{X} _{test}$ and $\mathcal{P}_u$ simulates two realistic scenes: 1. A group of closely interacting proteins, which are often distributed in clusters in the PPI network; 2. A group of proteins drifting away from the PPI network.
We select a root node $p_{\mathrm{root}}$, fix the size of the testset $N$, and then use the Breadth-First Search (BFS) algorithm in the PPI network to obtain the proteins $\mathcal{P}_u$ that meet the scenario 1. All PPIs related to these proteins are the generated testset. For scenario 2, we just need to simply randomly select proteins to form $\mathcal{P}_u$. However, in order to maintain the PPI network connectivity of $\mathcal{X} _{\mathrm{train}}$ and $\mathcal{X} _{\mathrm{test}}$, we use the Depth-First Search (DFS) algorithm to simulate. The details of the data partition algorithm are shown in Algorithm \ref{alg:algorithm}, where we will not show the details of the BFS and DFS algorithms but use the $\mathrm{Search\_Next}$ function to return the next protein of the current protein $p_{\mathrm{cur}}$ in different search algorithms. The $\mathcal{N}(p)$ returns all neighbors of protein $p$. And we controls the degree of the root node $|\mathcal{N}(p_{\mathrm{root}})| < t$ to simulate newly discovered proteins (Usually few proteins interact with them).

\begin{algorithm}[tb]
\caption{Data Partition Algorithm}
\label{alg:algorithm}
\textbf{Input}: Protein set $\mathcal{P}$; PPI set $\mathcal{X}$; Testset size $N$; Root node selection threshold ${t}$; Search order $\mathcal{S} \in \{\mathrm{BFS}, \mathrm{DFS}\}$;\\
\textbf{Output}: $\mathcal{X} _{\mathrm{train}}$; $\mathcal{X} _{\mathrm{test}}$;
\begin{algorithmic}[1] %[1] enables line numbers
\State Build PPI graph $\mathcal{G}=(\mathcal{P}, \mathcal{X})$ 
\LineComment{Root node selection}
%\STATE Randomly select a protein as root node $p _{root}$
\Repeat
    \State Randomly select a protein as root node $p _{\mathrm{root}}$.
\Until{$|\mathcal{N}(p _{\mathrm{root}})| < {t}$} \Comment{$\mathcal{N}$ returns the neighbors}
\LineComment{Testset construction}
\State $\mathcal{X} _{\mathrm{test}} = \emptyset$, $p _{\mathrm{cur}} = p _{\mathrm{root}}$
\Repeat
    \State $\mathcal{X} _{\mathrm{cur}} = \{ \{p _{\mathrm{cur}}, p_k\} | p_k \in \mathcal{N}(p _{\mathrm{cur}})\}$
    \State $\mathcal{X} _{\mathrm{test}} = \mathcal{X} _{\mathrm{test}} \cup \mathcal{X} _{\mathrm{cur}}$
    \State $p _{\mathrm{cur}} = \mathrm{Search\_Next}(\mathcal{G}, p _{\mathrm{cur}}, \mathcal{S})$
\Until{$|\mathcal{X} _{\mathrm{test}}| \geq N$}
\LineComment{Trainset construction}
\State $\mathcal{X} _{\mathrm{train}} = \mathcal{X} - \mathcal{X} _{\mathrm{test}}$
\State \textbf{return} $\mathcal{X} _{\mathrm{train}}, \mathcal{X} _{\mathrm{test}}$
\end{algorithmic}
\end{algorithm}

\subsection{Protein Feature Encoding}
% The Protein feature encoding of GNN-PPI is divided into two parts. First, the protein feature is encoded independently based on the amino acid sequence, and then the proteins are linked with other proteins according to the PPI network graph to capture the topological information of its neighbors for encoding.
% \subsubsection{Protein-Independent Feature} 

% The Protein-Independent Encoding(PIE) modules seek to leverage both the global sequential information and local features to generate more effective protein feature representations as to the PPI network input. This neural encoder contains the following instances, 1D-convolution layer(Conv1d) with pooling, bidirectional gated recurrent unit(BiGRU), and fully connected layer. The architecture of PIE is shown in Figure \ref{fig:picture001}. 1D-convolution layer with pooling is designed to adaptively encode local features of amino acid sequences, where Conv1d applies to combines the local feature from each $k$-mer of amino acid, and pooling preserves the most significant features and reduce computing consumption. Then we use BiGRU to characterizes the sequential information from two directions. Finally, GAP and FC encoders are used to obtain a one-dimensional sequence of the protein-independent feature.
Previous work \cite{chen2019multifaceted} has proved that protein features based on the amino acid sequence are beneficial to the performance improvement of PPI-related tasks. Therefore, we design a Protein-Independent Encoding (PIE) module, which contains Conv1d with pooling, BiGRU, and fully connected (FC) layer, to generate protein feature representations as input to the PPI network.

The subsequent Protein-Graph Encoding (PGE) module is the core of GNN-PPI. Inspired by PPI network being widely used in bioinformatics computing, we construct PPI network $\mathcal{G} = (\mathcal{P}, \mathcal{X})$, and convert the original independent learning tasks $\mathcal{F}(x_{ij}|p_i, p_j, \theta) \to \hat{y}_{ij}$ into graph-related learning tasks $\mathcal{F}(x_{ij}|\mathcal{G}, \theta) \to \hat{y}_{ij}$. Recently, GNN is the most effective graph representation learning method, its main idea is the recursive neighborhood aggregation scheme, where each node computes a new feature by aggregating the previous features of its neighbor nodes. After $k$ iterations, a node is represented by its transformed feature, which captures the structural information within the node’s $k$-hop neighborhood. More specifically, the GNN of the k-th iteration is $$ a^k_{p} = \mathrm{Agg}\Big(\{g^{k-1}_{p'} | p' \in \mathcal{N}(p)\}\Big), g^k_{p} = \mathrm{Update}\Big(\{ g^{k-1}_p, a^k_p\}\Big)$$ where $g^k_p$ is the feature of node $p$ at the k-th iteration. The design of Agg($\cdot$) and Update($\cdot$) are the keys to different GNN architectures. In this paper we use Graph Isomorphism Network (GIN) \cite{xu2018powerful},  where the sum of the neighbor node features is used as the aggregation function, and multi-layer perceptrons (MLPs) is used to update the aggregated features. Then, GIN updates node features as $$ g^k_p = \mathrm{MLP}^k \Big((1+\epsilon^k) \cdot g^{k-1}_p + \sum_{p' \in \mathcal{N}(p)} g^{k-1}_{p'} \Big) $$where $\epsilon$ can be a learnable parameter or a fixed scalar.

\subsection{Multi-Label PPI Prediction}
\label{section:3.5}

With the feature of protein learned from the previous stages for the PPI $x_{ij}$, we use the dot product operation to combine the features of $p_i$ and $p_j$, and then use a fully connected layer (FC) as classifier for multi-label PPI prediction, expressed as $ \hat{y}_{ij} = \mathrm{FC}(g_{p_i} \cdot g_{p_j}) $. The PIE and PGE modules are jointly training in an end-to-end way. Given a training set $\mathcal{X}_{\mathrm{train}}$ and its ground-truth multi-label interaction $\mathcal{Y}_{\mathrm{train}}$, we can use the multi-task binary cross-entropy as the loss function: $$ L = \sum_{k=0}^{n}\bigg( \sum_{x_{ij} \in \mathcal{X} _{\mathrm{train}}} -y_{ij}^k \log \hat{y}_{ij}^k - (1-y_{ij}^k) \log(1-\hat{y}_{ij}^k) \bigg).$$ 

Different from the algorithm that considers PPI independently, GNN-PPI learns to combine protein neighbors to generate feature representations. Therefore, for the $\mathcal{X} _{\mathrm{test}}$ constructed by our proposed BFS or DFS, GNN-PPI can also be based on its neighbors to generate suitable feature representations for multi-type PPI prediction. On the other hand, even if the PPI network $\mathcal{G}' = (\mathcal{P}_v, \mathcal{X} _{\mathrm{train}}) $ used in the training process is constructed with only $\mathcal{X} _{\mathrm{train}}$, it can perform well for unknown PPI $x_{ij} \in \mathcal{X} _{\mathrm{test}}$. (See details in Table \ref{tab:4})

\begin{table*}[htb]
\small
\centering
\begin{tabular}{ccccccccc}
\hline
\multirow{2}*{Dataset} & Partition & \multicolumn{7}{c}{Methods}\\
\cline{3-9}
    & Scheme & SVM  & RF & LR & DPPI & DNN-PPI & PIPR & GNN-PPI\\
\hline
\multirow{3}*{SHS27k} & Random  & 75.35$\pm$1.05  & 78.45$\pm$0.88 & 71.55$\pm$0.93                         & 73.99$\pm$5.04 & 77.89$\pm$4.97 & 83.31$\pm$0.75 & \textbf{87.91}$\pm$0.39\\

                      & BFS  & 42.98$\pm$6.15  & 37.67$\pm$1.57 & 43.06$\pm$5.05 
       & 41.43$\pm$0.56 & 48.90$\pm$7.24 & 44.48$\pm$4.44 & \textbf{63.81}$\pm$1.79\\

                      & DFS  & 53.07$\pm$5.16  & 35.55$\pm$2.22 & 48.51$\pm$1.87 
       & 46.12$\pm$3.02 & 54.34$\pm$1.30 & 57.80$\pm$3.24 & \textbf{74.72}$\pm$5.26\\
\hline
\multirow{3}*{SHS148k} & Random  & 80.55$\pm$0.23  & 82.10$\pm$0.20 & 67.00$\pm$0.07                        & 77.48$\pm$1.39 & 88.49$\pm$0.48 & 90.05$\pm$2.59 & \textbf{92.26}$\pm$0.10\\

                      & BFS  & 49.14$\pm$5.30  & 38.96$\pm$1.94 & 47.45$\pm$1.42 
       & 52.12$\pm$8.70 & 57.40$\pm$9.10 & 61.83$\pm$10.23 & \textbf{71.37}$\pm$5.33\\

                      & DFS  & 58.59$\pm$0.07  & 43.26$\pm$3.43 & 51.09$\pm$2.09 
       & 52.03$\pm$1.18 & 58.42$\pm$2.05 & 63.98$\pm$0.76 & \textbf{82.67}$\pm$0.85\\
\hline
\multirow{3}*{STRING} & Random  & -  & 88.91$\pm$0.08 & 67.74$\pm$0.16
       & 94.85$\pm$0.13 & 83.08$\pm$0.11 & 94.43$\pm$0.10 & \textbf{95.43}$\pm$0.10\\

                      & BFS  & -  & 55.31$\pm$1.02 & 50.54$\pm$2.00 
       & 56.68$\pm$1.04 & 53.05$\pm$0.82 & 55.65$\pm$1.60 & \textbf{78.37}$\pm$5.40\\

                      & DFS  & -  & 70.80$\pm$0.45 & 61.28$\pm$0.53 
       & 66.82$\pm$0.29 & 64.94$\pm$0.93 & 67.45$\pm$0.34 & \textbf{91.07}$\pm$0.58\\
\hline
\end{tabular}
\caption{Performance of GNN-PPI against comparative methods over different datasets and data partition schemes.  The reported results are mean$\pm$std micro-averaged F1 score over three repeated experiments. Results of SVM on STRING is omitted for unafforable running time.}
\label{tab:2}
\end{table*}

\section{Experiment}

\subsection{Dataset}

We use multi-type PPI data from the STRING database\footnote{https://string-db.org/} \cite{szklarczyk2019string} to evaluate our proposed GNN-PPI. The STRING database collected, scored, and integrated most publicly available sources of protein-protein interaction information and built a comprehensive and objective global PPI network, including direct (physical) and indirect (functional) interactions. In this paper, we focus on the multi-type classification of PPI by STRING. It divides PPI into 7 types, namely reaction, binding, post-translational modifications (ptmod), activation, inhibition, catalysis, and expression. Each pair of interacting proteins contains at least one of them. \cite{chen2019multifaceted} randomly select 1,690 and 5,189 proteins from the Homo sapiens subset of STRING that shares $<40\%$ of sequence identity to generate two subsets, namely SHS27k and SHS148k, which contain 7,624 and 44,488 multi-label PPIs. At the same time, we use all PPIs of Homo sapiens as our third dataset, namely STRING, which contains 15,335 proteins and 593,397 PPIs. We will use these three PPI datasets of different sizes to evaluate GNN-PPI and other PPI methods in the following content.

\subsection{Experimental Details}

% \subsubsection{Pre-train amino acid embeddings}

% To obtain protein input features based on amino acid sequences, we use the embedding method to represent each amino acid $a\in \mathcal{A}$ as a vector. Each embedding vector is a concatenation of two sub-embedding, i.e. $E(a)=[E_1(a), E_2(a)]$. The first part $E_1$ measures the co-occurrence similarity of the amino acids, which is obtained by pre-training the Skip-Gram\cite{mikolov2013distributed} model on protein sequences. The skip-gram model is trained using negative sampling, where the vocabulary samples are overlapping 3-mer amino acids, and the word vector size is 5. The second part $E_2$ is a one-hot encoding based on the classification defined by the similarity of electrostaticity and hydrophobicity among amino acids, where 22 natural amino acids can be clustered into 7 classes\cite{shen2007predicting}. In summary, each amino acid is expressed as $E(a) \in R^{5+(7+1)=13}$
% For the $21^{st}$ amino acid U(Selenocysteine), the $22^{nd}$ amino acid O(Pyrrolysine) and the unknown amino acids X are included in the eighth category. 

\subsubsection{Experimental Settings and Metrics}
\label{section:4.2}

We select 20\% of PPIs for testing, using our proposed BFS, DFS, and original evaluation (Random). The BFS or DFS partition algorithm has completely different results for different root nodes. To simulate the realistic scene mentioned in Section 3.3, the root node's degree should not be too large. We set the root node degree threshold $t=5$. To eliminate the influence of the randomness of data partitioning on the performance of PPI methods, we repeat experimental results under 3 different random seeds. We use the protein features based on amino acid sequence, refer to \cite{chen2019multifaceted} using embedding method to represent each amino acid (Details in the Appendix C). We adopt Adam algorithm \cite{kingma2014adam} to optimize all trainable parameters. The other hyper-parameters settings are shown in Appendix Table \ref{tab:55}.

We evaluate the multi-label PPI prediction performance using micro-F1. This is because micro-averaging will emphasize the common labels in the dataset, which gives each sample the same importance. Since the different PPI types in the datasets we used are very imbalanced, micro-F1 may be preferred. Even so, we still evaluate the F1 performance of each PPI type, and the results are shown in Table \ref{tab:66}.

\subsubsection{Baselines}

% GNN-PPI is compared with the following two types of PPI prediction models.

% \textbf{ML-Based:} SVM\cite{guo2008using}, RF\cite{wong2015detection}, and LR\cite{silberberg2014method}. The input feature of the algorithm uniformly selects common protein features by artificial design, AC and CTD, of which CTD use seven attributes for the division.

% \textbf{DL-Based:} DPPI\cite{hashemifar2018predicting}, DNN-PPI\cite{li2018deep}, and PIPR\cite{chen2019multifaceted}. We construct the same architecture as the original paper and modify the output of the original implementation from a binary class to multi-label. The protein input feature based on the amino acid sequence is consistent with GNN-PPI. 

We compare GNN-PPI against a variety of baselines, which can be categorized as follows:

1. \textbf{Machine Learning based:} We choose three representative machine learning (ML) algorithms, SVM \cite{guo2008using}, RF \cite{wong2015detection}, and LR \cite{silberberg2014method}. The input feature of the algorithms uniformly selects common handcrafted protein features, AC \cite{guo2008using} and CTD \cite{du2017deepppi}, of which CTD use seven attributes for the division (See in Appendix B).

2. \textbf{Deep Learning based:} We choose three representative deep learning (DL) algorithms in PPI prediction, PIPR \cite{chen2019multifaceted}, DNN-PPI \cite{li2018deep} and DPPI \cite{hashemifar2018predicting}. We construct the same architecture as the original papers and modify the output of the original implementation from a binary class to multi-label. The protein input feature based on the amino acid sequence is consistent with GNN-PPI. The other settings are the same as the original papers.

\begin{table*}[htb]
\small
\centering
\begin{tabular}{cccc|cc|cc|R{0.7cm}R{0.7cm}R{0.7cm}cc}
\hline
\multirow{3}*{Dataset} & Partition & \multicolumn{2}{c}{$\mathcal{X} _{\mathrm{BS}}$} & \multicolumn{2}{c}{$\mathcal{X} _{\mathrm{ES}}$} & \multicolumn{2}{c}{$\mathcal{X} _{\mathrm{NS}}$} & \multicolumn{5}{c}{$\mathcal{X}_{\mathrm{Avg}}$} \\
\cmidrule(r){3-4} \cmidrule(r){5-6} \cmidrule(r){7-8} \cmidrule(r){9-13}
 & Scheme & PIPR & GNN-PPI & PIPR & GNN-PPI & PIPR & GNN-PPI & \multicolumn{3}{c}{Proportion(BS/ES/NS)} & PIPR & GNN-PPI\\ 
\hline
\multirow{3}{*}{SHS27k} & Random & 83.12 & 88.31 & 64.48 & 74.28 & 35.29 & 33.33 & 92.2 & 7.5 & 0.3 & 81.58 & 87.11\\
                        & BFS    & -     & -     & 44.92 & 68.08 & 30.34 & 46.25 & 0.0 & 72.6 & 27.4 & 40.92 & 62.10\\
                        & DFS    & -     & -     & 58.25 & 72.22 & 48.77 & 63.22 & 0.0 & 88.6 & 11.4 & 57.17 & 71.19\\
\hline
\multirow{3}{*}{SHS148k} & Random & 92.82 & 92.24 & 78.80 & 73.09 & 40.72 & 36.36 & 97.2 & 2.7 & 0.1 & 92.42 & 91.68\\
                        & BFS    & -     & -     & 62.80 & 72.51 & 73.82 & 77.02 & 0.0 & 69.7 & 30.3 & 66.13 & 73.88\\
                        & DFS    & -     & -     & 64.17 & 83.37 & 55.51 & 73.08 & 0.0 & 91.9 & 8.1 & 63.47 & 82.54\\
\hline
\multirow{3}{*}{STRING} & Random & 94.32 & 95.42 & 61.65 & 77.68 & 33.33 & 57.14 & 99.7 & 0.3 & 0 & 94.23 & 95.37\\
                        & BFS    & -     & -     & 56.71 & 83.99 & 39.87 & 72.83 & 0.0 & 85.8 & 14.2 & 54.31 & 82.41\\
                        & DFS    & -     & -     & 68.61 & 90.38 & 55.22 & 87.07 & 0.0 & 94.3 & 5.7 & 67.84 & 90.19\\
\hline
\end{tabular}
\caption{In-depth analysis between PIPR and GNN-PPI over BS, ES and NS subsets.}
\label{tab:3}
\end{table*}

\subsection{Results and Analysis}

\subsubsection{Benchmark}

Table \ref{tab:2} compares the performance of different methods under different evaluations and different datasets. Firstly, consider the impact of different evaluations, we can see that any method in Table \ref{tab:2} perform well under Random partition. However, under BFS or DFS partition, except for GNN-PPI, the performance of other methods declines clearly. Moreover, the performance under the DFS is generally higher than that of the BFS, which means that the clustered distribution of unknown proteins in the PPI network is harder to learn than discrete distribution.
% e.g. in the SHS27k dataset, the micro-F1 of PIPR drops to 44.48 in BFS partition, which is equivalent to random guessing.
Next, observe the performance on different datasets. Regardless of the evaluations, the performance of any method will improve as the data size increases. However, the problems mentioned above will not be trivially solved by increasing the amount of data.
Finally, comparing different methods, we can see that DL-based methods are generally better than ML-based, and GNN-PPI can achieve state-of-the-art performance. However, under the Random partition, the advantage of GNN-PPI over DL-based methods will be smaller as the dataset size increases. The most prominent advantage of GNN-PPI is that under the BFS or DFS partition, and for the inter-novel-protein interactions, it can still learn useful feature representations from protein neighbors so as to obtain good performance in  PPI prediction.
In summary, the experimental results show that GNN-PPI can effectively improve the prediction accuracy of inter-novel-protein interactions. However, how to further push the performance to be comparable as Random partition is still a problem worthy of further discussion, and it is also our future work.

\subsubsection{In-depth Analysis}

We make a more in-depth analysis of performance between PIPR and GNN-PPI on $\mathcal{X} _{\mathrm{test}}$, as shown in Table \ref{tab:3}. Observing the proportions of different subsets of the testset, we can find that under Random partition, more than 92\% test samples belong to $\mathcal{X} _{\mathrm{BS}}$, which is consistent with our corollary \ref{corollary:1}. PIPR performs well on the randomly divided testset (81.58 in SHS27k, 92.42 in SHS148k, and 94.23 in STRING), but if we further investigate the testset, we will find that PIPR performs very poorly for inter-novel-protein interactions ($x_{ij} \in \mathcal{X} _{\mathrm{ES}}$ or $\mathcal{X} _{\mathrm{NS}}$), but it is dominated by $\mathcal{X} _{\mathrm{BS}}$, which has accurate performance and a high proportion. According to the results of Table \ref{tab:2} and Table \ref{tab:3}, with sufficient $\mathcal{X} _{\mathrm{ES}}$ and $\mathcal{X} _{\mathrm{NS}}$ data, we can assert that the methods which treats PPI as an independent sample (represented by PIPR), cannot accurately predict inter-novel-protein interactions. On the contrary, our proposed GNN-PPI can still perform well under BFS and DFS. Moreover, as the data size increases, the performance of GNN-PPI is better (e.g., 82.41 vs. 54.31 in STRING-BFS and 90.19 vs. 67.84 in STRING-DFS).

\begin{table}[tb]
\small
\centering
\begin{tabular}{p{1.1cm}<{\centering}|p{1.3cm}<{\centering}p{1.45cm}<{\centering}ccc}
\hline
\multirow{2}*{Methods} & \multirow{2}*{Trainset} & \multirow{2}*{Testset} & \multicolumn{3}{c}{Partition Scheme} \\
\cline{4-6}
 & & & Random & BFS & DFS\\
\hline
\multirow{4}{*}{PIPR} & \multirow{2}{*}{\scriptsize{SHS27k-Train}}& \scriptsize{SHS27k-Test} & 81.58 & 40.92 & 57.17\\
          & & \scriptsize{STRING} & 42.79 & 48.55 & 57.44\\
\cline{2-6}
                    & \multirow{2}{*}{\scriptsize{SHS148k-Train}}& \scriptsize{SHS148k-Test} & 92.42 & 66.13 & 63.47\\
       & & \scriptsize{STRING} & 53.85 & 63.74 & 62.46\\
\hline
\multirow{4}{*}{GNN-PPI} & \multirow{2}{*}{\scriptsize{SHS27k-Train}} & \scriptsize{SHS27k-Test} & 87.11 & 62.10 & 71.19\\
           & & \scriptsize{STRING} & 66.85 & 66.39 & 67.43\\
\cline{2-6}
                         & \multirow{2}{*}{\scriptsize{SHS148k-Train}} & \scriptsize{SHS148k-Test} & 91.68 & 73.88 & 82.54\\
           & & \scriptsize{STRING} & 73.12 & 67.43 & 70.64\\
\hline
\end{tabular}
\caption{Performance comparison of tested on trainset-homologous testset vs. unseen testset, under different evaluations (partition schemes).}
\label{tab:5}
\end{table}

\subsubsection{Model Generalization}

We study the ability of different evaluations to assess the model's generalization. We take the trained model's test performance on the larger dataset STRING as the model's true generalization ability. If the gap between the trainset-homologous test performance and the generalization is smaller, then the evaluation can better reflect the model's generalization. The experimental results are shown in Table \ref{tab:5}. It can be seen that the previous evaluation (Random), whether it is for PIPR or GNN-PPI, the test performance on the STRING dataset has severely dropped. Like our speculation, it cannot reflect the generalization of the model. On the contrary, under the evaluation of BFS or DFS, its test performance can truly reflect the performance of the model, no matter it is good or bad (e.g. 66.13 vs. 63.74 in PIPR-SHS148k-BFS and 71.19 vs. 67.43 in GNN-PPI-SHS27k-DFS). In fact, the testset obtained by BFS or DFS is theoretically the same as the sample tested on STRING. The only difference is the proportion of different types of PPI (BS, ES and NS). Testing On the STRING, the proportion of NS is higher.

\begin{table}[tb]
\small
\centering
\begin{tabular}{ccccc}
\hline
Partition & \multirow{2}*{Graph} & \multicolumn{3}{c}{Dataset} \\
\cline{3-5}
Scheme & & SHS27k & SHS148k & STRING\\
\hline
\multirow{2}*{BFS} & GCA & 63.81$\pm$1.79 & 71.37$\pm$5.33 & 78.37$\pm$5.40 \\
                   & GCT & 60.61$\pm$5.32 & 69.56$\pm$6.89 & 73.23$\pm$3.93 \\
\hline
\multirow{2}*{DFS} & GCA & 74.72$\pm$5.26 & 82.67$\pm$0.85 & 91.07$\pm$0.58 \\
                   & GCT & 73.42$\pm$5.50 & 80.35$\pm$2.20 & 89.04$\pm$1.06 \\
\hline
\end{tabular}
\caption{Performance of GNN-PPI with different PPI Graph construction method.}
\label{tab:4}
\end{table}

% \subsection{Ablation study}
% \label{section:4.4}

\subsubsection{PPI Network Graph Construction}

We study the impact of the PPI network graph construction method (mentioned in \ref{section:3.5}) in the GNN-PPI. There are two graph construction methods, graph construct by all data (GCA, $\mathcal{G}=(\mathcal{P}, \mathcal{X})$) and graph construct by trainset (GCT, $\mathcal{G}=(\mathcal{P}_v, \mathcal{X} _{\mathrm{train}})$). The experimental results are shown in Table \ref{tab:4}. It can be seen that the performance of GCA all exceeds that of GCT, which is reasonable because the graph construction of GCA accesses more complete information than GCT. Compared with BFS, in the case of DFS, the performance of GCT is closer to GCA, which seems to indicate that the protein neighbors are more complete, the performance will be better. What is more noteworthy is that GCT is still much higher than non-graph algorithms, which shows the superiority of GNN in processing the few-shot learning for multi-label PPI prediction task. Moreover, for unknown proteins, we often cannot know their neighbors in advance. The effectiveness of GCT shows that the trained model is robust to newly discovered proteins and their interactions.

% \subsection{Supplementary Experiment}

\subsubsection{Performance of Different PPI Types}

\begin{table*}[tb]
\centering
\begin{tabular}{cccccccc}
\hline
\multirow{2}*{Multi Labels} & \multirow{2}*{Type Ratio (\%)} & \multicolumn{2}{c}{Random Partition} & \multicolumn{2}{c}{BFS Partition} & \multicolumn{2}{c}{DFS Partition} \\
 \cmidrule(lr){3-4} \cmidrule(lr){5-6} \cmidrule(lr){7-8}
  &  & PIPR & GNN-PPI & PIPR & GNN-PPI & PIPR & GNN-PPI\\
\hline
Reaction & 51.08  & 96.39$\pm$0.12 & 97.62$\pm$0.07 & 55.96$\pm$7.59 & 83.19$\pm$4.01 & 68.09$\pm$1.19 & 93.28$\pm$1.44 \\
Binding  & 67.87   & 95.63$\pm$0.23 & 96.43$\pm$0.07 & 71.34$\pm$0.36 & 83.80$\pm$3.70 & 81.79$\pm$1.90 & 94.06$\pm$0.51 \\
Ptmod    & 6.92   & 86.94$\pm$0.26 & 87.28$\pm$0.27 & 25.91$\pm$13.8 & 70.48$\pm$7.44 & 17.09$\pm$5.31 & 82.12$\pm$1.40 \\
Activation & 17.53 & 86.31$\pm$0.31 & 87.96$\pm$0.37 & 39.17$\pm$13.3 & 66.20$\pm$15.5 & 27.28$\pm$10.0 & 81.58$\pm$0.94 \\
Inhibition & 6.58 & 90.36$\pm$0.34 & 91.49$\pm$0.14 & 12.08$\pm$7.55 & 65.58$\pm$12.4 & 19.16$\pm$0.94 & 82.62$\pm$2.80 \\
Catalysis  & 48.59 & 96.28$\pm$0.19 & 97.58$\pm$0.08 & 58.84$\pm$6.39 & 83.94$\pm$4.53 & 66.24$\pm$4.32 & 92.79$\pm$0.51 \\
Expression & 2.07 & 39.06$\pm$1.42 & 32.55$\pm$1.53 & 0.81$\pm$1.27 & 15.67$\pm$10.8 & 1.04$\pm$1.80 & 23.22$\pm$9.21 \\
\hline
Micro-Avg   & -  & 94.43$\pm$0.10 & 95.43$\pm$0.10 & 55.65$\pm$1.60 & 78.37$\pm$5.40 & 67.45$\pm$0.34 & 91.07$\pm$0.58 \\
\hline
\end{tabular}
\caption{Separate results in STRING dataset for the multi labels between PIPR and GNN-PPI over Random, BFS and DFS partition schemes.}
\label{tab:66}
\end{table*}

Table \ref{tab:66} shows the proportions of different PPI types and their performance. It can be seen that although there are many types of PPI, the types are relatively unbalanced, and the proportions of Ptmod, Inhibition, and Expression are less than 10\%. As mentioned in Section \ref{section:4.2}, it is unreasonable to use micro-F1 for evaluation, but in supplementary experiments, we still show its performance truthfully. Under the random partition scheme, whether it is PIPR or GNN-PPI, the performance of different PPI types is good, except for the Expression type, only 39.06 and 32.55. However, under the BFS or DFS partition schemes, PIPR cannot predict the Inter-novel-protein Interaction, and there will be a relatively large performance degradation, and if the proportion of PPI type ratio is lower, the performance degradation will be more serious. Among them, the Inhibition type drops from 90.36 to 12.08 in BFS schemes, and the expression type completely unpredictable, the performance is only 0.81. Our model GNN-PPI demonstrates consistent advantage over PIPR among all the 7 labels. But it is undeniable that the performance of the PPI type with a low proportion is still poor, which is also one of our future works.

\subsubsection{Ablation Study of PIE and PGE}

Table \ref{tab:44} shows ablation studies on the PIE and PGE components. It is worth mentioning that when the model has only the PGE module, it is limited by memory, and the input features of the model select common handcrafted protein features, AC and CTD (as mentioned in Section \ref{section:4.2}). When the model has only the PIE module, the model degenerates to Deep Learning based methods, and its performance changes in Random, BFS and DFS are the same as PIPR, DNN-PPI and DPPI. When there is only the PGE module, the performance advantages of the graph model algorithm are revealed, and the Inter-novel-protein Interaction still has strong performance. In general, as proved by previous work \cite{chen2019multifaceted}, it is better to use amino acid sequence based protein features than common handcrafted protein features, PIE and PGE modules are beneficial to the overall performance.

\begin{table}[tb]
\small
\centering
\begin{tabular}{ccccc}
\hline
\multirow{2}*{PIE} & \multirow{2}*{PGE} & \multicolumn{3}{c}{Partition Schemes}\\
\cline{3-5}
    & & Random & BFS & DFS\\
\hline
\checkmark &    & 69.88$\pm$0.04 & 50.03$\pm$2.08 & 61.86$\pm$1.04 \\
 & \checkmark   & 94.30$\pm$0.52 & 73.81$\pm$6.82 & 88.03$\pm$0.59 \\
\checkmark & \checkmark & 95.38$\pm$0.12 & 78.37$\pm$5.40 & 91.07$\pm$0.58 \\
\hline
\end{tabular}
\caption{Results of ablation studies on the PIE and PGE components.}
\label{tab:44}
\end{table}

\section{Conclusion}

In this paper, we study the significant performance degradation of existing PPI methods when tested in unseen dataset. Experimental results show that this problem is due to the poor performance of the model for inter-novel-protein interactions. However, current evaluation overlook the inter-novel-protein interactions, and are thus not instructive for the performance when tested on unseen datasets. Therefore, we design a new evaluation framework with two per-protein randomized data partition startegies, namely BFS and DFS, and propose a GNN based method GNN-PPI to model the correlations between PPIs. Our experimental results show that GNN-PPI outperforms state-of-the-art PPI prediction methods regardless of the evaluation is original or our proposed, especially for the inter-novel-protein interactions prediction.

\bibliographystyle{named}
\bibliography{ijcai21}

\begin{thebibliography}{}

\bibitem[\protect\citeauthoryear{Anfinsen}{1972}]{anfinsen1972formation}
Christian~B Anfinsen.
\newblock The formation and stabilization of protein structure.
\newblock {\em Biochemical Journal}, 128(4):737--749, 1972.

\bibitem[\protect\citeauthoryear{Chen \bgroup \em et al.\egroup
  }{2019}]{chen2019multifaceted}
Muhao Chen, Chelsea J-T Ju, Guangyu Zhou, Xuelu Chen, Tianran Zhang, Kai-Wei
  Chang, Carlo Zaniolo, and Wei Wang.
\newblock Multifaceted protein--protein interaction prediction based on siamese
  residual rcnn.
\newblock {\em Bioinformatics}, 35(14):i305--i314, 2019.

\bibitem[\protect\citeauthoryear{De~Las~Rivas and
  Fontanillo}{2010}]{de2010protein}
Javier De~Las~Rivas and Celia Fontanillo.
\newblock Protein--protein interactions essentials: key concepts to building
  and analyzing interactome networks.
\newblock {\em PLoS Comput Biol}, 6(6):e1000807, 2010.

\bibitem[\protect\citeauthoryear{Deng \bgroup \em et al.\egroup
  }{2020}]{deng2020multimodal}
Yifan Deng, Xinran Xu, Yang Qiu, Jingbo Xia, Wen Zhang, and Shichao Liu.
\newblock A multimodal deep learning framework for predicting drug-drug
  interaction events.
\newblock {\em Bioinformatics}, 2020.

\bibitem[\protect\citeauthoryear{Du \bgroup \em et al.\egroup
  }{2017}]{du2017deepppi}
Xiuquan Du, Shiwei Sun, Changlin Hu, Yu~Yao, Yuanting Yan, and Yanping Zhang.
\newblock Deepppi: boosting prediction of protein--protein interactions with
  deep neural networks.
\newblock {\em Journal of chemical information and modeling}, 57(6):1499--1510,
  2017.

\bibitem[\protect\citeauthoryear{Dubchak \bgroup \em et al.\egroup
  }{1995}]{dubchak1995prediction}
Inna Dubchak, Ilya Muchnik, Stephen~R Holbrook, and Sung-Hou Kim.
\newblock Prediction of protein folding class using global description of amino
  acid sequence.
\newblock {\em Proceedings of the National Academy of Sciences},
  92(19):8700--8704, 1995.

\bibitem[\protect\citeauthoryear{ERDdS and R\&wi}{1959}]{erdds1959random}
P~ERDdS and A~R\&wi.
\newblock On random graphs i.
\newblock {\em Publ. math. debrecen}, 6(290-297):18, 1959.

\bibitem[\protect\citeauthoryear{Erd{\H{o}}s and
  R{\'e}nyi}{1960}]{erdHos1960evolution}
Paul Erd{\H{o}}s and Alfr{\'e}d R{\'e}nyi.
\newblock On the evolution of random graphs.
\newblock {\em Publ. Math. Inst. Hung. Acad. Sci}, 5(1):17--60, 1960.

\bibitem[\protect\citeauthoryear{Fields and Song}{1989}]{fields1989novel}
Stanley Fields and Ok-kyu Song.
\newblock A novel genetic system to detect protein--protein interactions.
\newblock {\em Nature}, 340(6230):245--246, 1989.

\bibitem[\protect\citeauthoryear{Guo \bgroup \em et al.\egroup
  }{2008}]{guo2008using}
Yanzhi Guo, Lezheng Yu, Zhining Wen, and Menglong Li.
\newblock Using support vector machine combined with auto covariance to predict
  protein--protein interactions from protein sequences.
\newblock {\em Nucleic acids research}, 36(9):3025--3030, 2008.

\bibitem[\protect\citeauthoryear{Hashemifar \bgroup \em et al.\egroup
  }{2018}]{hashemifar2018predicting}
Somaye Hashemifar, Behnam Neyshabur, Aly~A Khan, and Jinbo Xu.
\newblock Predicting protein--protein interactions through sequence-based deep
  learning.
\newblock {\em Bioinformatics}, 34(17):i802--i810, 2018.

\bibitem[\protect\citeauthoryear{Kingma and Ba}{2014}]{kingma2014adam}
Diederik~P Kingma and Jimmy Ba.
\newblock Adam: A method for stochastic optimization.
\newblock {\em arXiv preprint arXiv:1412.6980}, 2014.

\bibitem[\protect\citeauthoryear{Kipf and Welling}{2016}]{kipf2016semi}
Thomas~N Kipf and Max Welling.
\newblock Semi-supervised classification with graph convolutional networks.
\newblock {\em arXiv preprint arXiv:1609.02907}, 2016.

\bibitem[\protect\citeauthoryear{Li \bgroup \em et al.\egroup
  }{2018}]{li2018deep}
Hang Li, Xiu-Jun Gong, Hua Yu, and Chang Zhou.
\newblock Deep neural network based predictions of protein interactions using
  primary sequences.
\newblock {\em Molecules}, 23(8):1923, 2018.

\bibitem[\protect\citeauthoryear{Mikolov \bgroup \em et al.\egroup
  }{2013}]{mikolov2013distributed}
Tomas Mikolov, Ilya Sutskever, Kai Chen, Greg~S Corrado, and Jeff Dean.
\newblock Distributed representations of words and phrases and their
  compositionality.
\newblock {\em Advances in neural information processing systems},
  26:3111--3119, 2013.

\bibitem[\protect\citeauthoryear{Nambiar \bgroup \em et al.\egroup
  }{2020}]{nambiar2020transforming}
Ananthan Nambiar, Maeve Heflin, Simon Liu, Sergei Maslov, Mark Hopkins, and
  Anna Ritz.
\newblock Transforming the language of life: Transformer neural networks for
  protein prediction tasks.
\newblock In {\em Proceedings of the 11th ACM International Conference on
  Bioinformatics, Computational Biology and Health Informatics}, pages 1--8,
  2020.

\bibitem[\protect\citeauthoryear{Saha and others}{2020}]{saha2020amalgamation}
Sriparna Saha et~al.
\newblock Amalgamation of protein sequence, structure and textual information
  for improving protein-protein interaction identification.
\newblock In {\em Proceedings of the 58th Annual Meeting of the Association for
  Computational Linguistics}, pages 6396--6407, 2020.

\bibitem[\protect\citeauthoryear{Shen \bgroup \em et al.\egroup
  }{2007}]{shen2007predicting}
Juwen Shen, Jian Zhang, Xiaomin Luo, Weiliang Zhu, Kunqian Yu, Kaixian Chen,
  Yixue Li, and Hualiang Jiang.
\newblock Predicting protein--protein interactions based only on sequences
  information.
\newblock {\em Proceedings of the National Academy of Sciences},
  104(11):4337--4341, 2007.

\bibitem[\protect\citeauthoryear{Silberberg \bgroup \em et al.\egroup
  }{2014}]{silberberg2014method}
Yael Silberberg, Martin Kupiec, and Roded Sharan.
\newblock A method for predicting protein-protein interaction types.
\newblock {\em PLoS One}, 9(3):e90904, 2014.

\bibitem[\protect\citeauthoryear{Skrabanek \bgroup \em et al.\egroup
  }{2008}]{skrabanek2008computational}
Lucy Skrabanek, Harpreet~K Saini, Gary~D Bader, and Anton~J Enright.
\newblock Computational prediction of protein--protein interactions.
\newblock {\em Molecular biotechnology}, 38(1):1--17, 2008.

\bibitem[\protect\citeauthoryear{Stark \bgroup \em et al.\egroup
  }{2006}]{stark2006biogrid}
Chris Stark, Bobby-Joe Breitkreutz, Teresa Reguly, Lorrie Boucher, Ashton
  Breitkreutz, and Mike Tyers.
\newblock Biogrid: a general repository for interaction datasets.
\newblock {\em Nucleic acids research}, 34(suppl\_1):D535--D539, 2006.

\bibitem[\protect\citeauthoryear{Sun \bgroup \em et al.\egroup
  }{2017}]{sun2017sequence}
Tanlin Sun, Bo~Zhou, Luhua Lai, and Jianfeng Pei.
\newblock Sequence-based prediction of protein protein interaction using a
  deep-learning algorithm.
\newblock {\em BMC bioinformatics}, 18(1):1--8, 2017.

\bibitem[\protect\citeauthoryear{Szklarczyk \bgroup \em et al.\egroup
  }{2016}]{szklarczyk2016string}
Damian Szklarczyk, John~H Morris, Helen Cook, Michael Kuhn, Stefan Wyder, Milan
  Simonovic, Alberto Santos, Nadezhda~T Doncheva, Alexander Roth, Peer Bork,
  et~al.
\newblock The string database in 2017: quality-controlled protein--protein
  association networks, made broadly accessible.
\newblock {\em Nucleic acids research}, page gkw937, 2016.

\bibitem[\protect\citeauthoryear{Szklarczyk \bgroup \em et al.\egroup
  }{2019}]{szklarczyk2019string}
Damian Szklarczyk, Annika~L Gable, David Lyon, Alexander Junge, Stefan Wyder,
  Jaime Huerta-Cepas, Milan Simonovic, Nadezhda~T Doncheva, John~H Morris, Peer
  Bork, et~al.
\newblock String v11: protein--protein association networks with increased
  coverage, supporting functional discovery in genome-wide experimental
  datasets.
\newblock {\em Nucleic acids research}, 47(D1):D607--D613, 2019.

\bibitem[\protect\citeauthoryear{Wong \bgroup \em et al.\egroup
  }{2015}]{wong2015detection}
Leon Wong, Zhu-Hong You, Shuai Li, Yu-An Huang, and Gang Liu.
\newblock Detection of protein-protein interactions from amino acid sequences
  using a rotation forest model with a novel pr-lpq descriptor.
\newblock In {\em International Conference on Intelligent Computing}, pages
  713--720. Springer, 2015.

\bibitem[\protect\citeauthoryear{Xu \bgroup \em et al.\egroup
  }{2018}]{xu2018powerful}
Keyulu Xu, Weihua Hu, Jure Leskovec, and Stefanie Jegelka.
\newblock How powerful are graph neural networks?
\newblock {\em arXiv preprint arXiv:1810.00826}, 2018.

\bibitem[\protect\citeauthoryear{Yang \bgroup \em et al.\egroup
  }{2020}]{yang2020graph}
Fang Yang, Kunjie Fan, Dandan Song, and Huakang Lin.
\newblock Graph-based prediction of protein-protein interactions with
  attributed signed graph embedding.
\newblock {\em BMC bioinformatics}, 21(1):1--16, 2020.

\end{thebibliography}

\begin{appendices}

\section{Corollary on Random Partition Strategy}

\subsection{ER Random Graph Model}
Before the proof of corollary, we first introduce the Erdős–Rényi(ER) random graph model\cite{erdds1959random} in graph theory. There are two closely related variants of the model, introduced as follows:
\begin{enumerate}
    \item In the $\mathcal{G}(n, M)$ model, a graph is chosen uniformly at random from the collection of all graphs which have $n$ nodes and $M$ edges.
    \item In the $\mathcal{G}(n, p)$ model, a graph is constructed by connecting nodes randomly. Each edge is included in the graph with probability p independent from every other edge. 
\end{enumerate}

The behavior of random graphs is often studied in the case where $n$, the number of nodes, tends to infinity. Although $p$ and $M$ can be fixed in this case, they can also be functions depending on $n$. 

\cite{erdHos1960evolution} described the behavior of $\mathcal{G}(n, p)$ very precisely for various values of $p$ when $n$ tends to infinity. Their results include the following lemma:
\begin{lemma}
\label{lemma:1}
If $p > \frac{\ln{n}}{n}$, then a graph in $\mathcal{G}(n, p)$ will almost surely be connected.
\end{lemma}

The expected number of edges in $\mathcal{G}(n, p)$ is $\mathrm{C}_n^2  p$, and by the law of large numbers any graph in $\mathcal{G}(n, p)$ will almost surely have approximately this many edges (provided the expected number of edges tends to infinity). Therefore, a rough heuristic is that if $pn^2 \to \infty$ then $\mathcal{G}(n,p)$ should behave similarly to $\mathcal{G}(n, M)$ with $M = \mathrm{C}_n^2 p$ as $n$ increases\cite{erdHos1960evolution}. Therefore, we can get the following lemma based on \textbf{Lemma \ref{lemma:1}}:
\begin{lemma}
\label{lemma:2}
If $M > \frac{(n-1)\ln{n}}{2}$, then a graph in $\mathcal{G}(n, M)$ will almost surely be connected.
\end{lemma}

\subsection{Random Partition Strategy in the PPI Network}
As mentioned in section 3.3 of the original paper, we propose a corollary as follows:
\begin{corollary}
\label{corollary:11}
Randomly divide the PPI dataset, select $t \leq 0.2$ as the test set, then most of the proteins in the test set were seen in training.
\end{corollary}

The above corollary is equivalent to whether the training set protein includes most of the protein in the dataset. Review our problem formulation in the original paper: Given the Protein set $\mathcal{P}$ and PPI set $\mathcal{X}$, where $|\mathcal{P}| = N, |\mathcal{X}| = M$. The PPI network is denoted as $\mathcal{G}=(\mathcal{P}, \mathcal{X})=(N, M)$, and assume $\mathcal{G}$ is connected (The $\mathcal{G}$ used in the original paper are all connected). After using the random data partition strategy, if the connectivity of the training PPI network $\mathcal{G}_{\mathrm{train}}=(\mathcal{P}, \mathcal{X}_{\mathrm{train}})$ is large, then the \textbf{corollary \ref{corollary:11}} will be proved. In the real-world PPI dataset, the number of proteins is not infinity. Therefore, we can roughly judge whether our \textbf{Corollary} \ref{corollary:11} is correct based on \textbf{Lemma} \ref{lemma:2}. The experimental results are shown in Table \ref{tab:11}. No matter theoretical deductions($|\mathcal{X}_{\mathrm{train}}| > M'$) or real test results(Proportion of $\mathcal{X}_{\mathrm{BS}}$), it shows our \textbf{Corollary} \ref{corollary:11} is right.

It is worth mentioning that regarding dataset SHS148k and STRING, why $|\mathcal{X}_{\mathrm{train}}| \gg M' $ but the proportion of $\mathcal{X}_{\mathrm{BS}}$ still does not reach the proportion of connected graphs, which is equal to 1. This is because there are many proteins in the PPI network, and they only interact with one protein.(Shown in the column $deg(p) = 1$ of Table \ref{tab:11})

\begin{table*}[htb]
\small
\centering
\begin{tabular}{cccccccccc}
\hline
\multirow{2}*{Dataset} & \multirow{2}*{$\mathcal{P}_v$} & \multirow{2}*{$\mathcal{P}_u$} & \multirow{2}*{$\mathcal{X}_{\mathrm{train}}$} & \multirow{2}*{$\mathcal{X}_{\mathrm{test}}$} & \multirow{2}*{$M'$} & \multicolumn{3}{c}{Proportion (\%)} & \multirow{2}*{$deg(p) = 1$}\\
\cline{7-9}
 & & & & & & $\mathcal{X} _{\mathrm{BS}}$ & $\mathcal{X} _{\mathrm{ES}}$ & $\mathcal{X} _{\mathrm{NS}}$ \\
\hline
SHS27k & 1587.6 & 102.3 & 6099 & 1525 & 6276.7 & 92.66 & 6.95 & 0.39 & 409\\
SHS148k & 4971 & 218 & 35590 & 8898 & 22189.8 & 97.25 & 2.72 & 0.03 & 1016\\
STRING & 15082.3 & 252.6 & 474717 & 118680 & 73893.7 & 99.75 & 0.25 & 0 & 1044\\
\hline
\end{tabular}
\caption{The details of the real-world PPI dataset we used under random partition strategy; $M'$ is equal to $\frac{(n-1)\ln{n}}{2}$, where $n = |\mathcal{P}_v| + |\mathcal{P}_u|$; $deg(p) = 1$ means there is noly one interaction related to protein $p$.}
\label{tab:11}
\end{table*}

\begin{table*}[htb]
\small
\centering
\begin{tabular}{ccccc}
\hline
No. & Property & Class1 & Class2 & Class3\\[4pt]
\multirow{2}{*}{1} & \multirow{2}{*}{Hydrophobicity} & Polar & Neutral & Hydrophobicity\\
 & & R,K,E,D,Q,N & G,A,S,T,P,H,Y & C,L,V,I,M,F,W\\[4pt]
\multirow{2}{*}{2} & Normalized van der & 0-2.78 & 2.95-4.0 & 4.03-8.08\\
 &  Waals volume & G,A,S,T,P,D & N,V,E,Q,I,L & M,H,K,F,R,Y,W\\[4pt]
\multirow{2}{*}{3} & \multirow{2}{*}{Polarity} & 4.9-6.2 & 8.0-9.2 & 10.4-13.0\\
 & & L,I,F,W,C,M,V,Y & P,A,T,G,S & H,Q,R,K,N,E,D\\[4pt]
\multirow{2}{*}{4} & \multirow{2}{*}{Charge} & Positive & Neutral & Negative\\
 & & K,R & A,N,C,Q,G,H,I,L,M,F,P,S,T,W,Y,V & D,E\\[4pt]
\multirow{2}{*}{5} & \multirow{2}{*}{Secondary Structure} & Helix & Strand & Coil\\
 & & E,A,L,M,Q,K,R,H & V,I,Y,C,W,F,T & G,N,P,S,D\\[4pt]
\multirow{2}{*}{6} & \multirow{2}{*}{Solvent Accessibility} & Buried & Exposed & Intermediate\\
 & & A,L,F,C,G,I,V,W & P,K,Q,E,N,D & M,P,S,T,H,Y\\[4pt]
\multirow{2}{*}{7} & \multirow{2}{*}{Polarizability} & 0-1.08 & 0.128-0.186 & 0.219-0.409\\
 & & G,A,S,D,T & C,P,N,V,E,Q,I,L & K,M,H,F,R,Y,W\\[4pt]
\hline
\end{tabular}
\caption{Seven attributes and the division of the amino acids.}
\label{tab:22}
\end{table*}

\section{Composition(C), Transition(T) and Distribution(D)}

\cite{dubchak1995prediction} proposes to use these attributes to describe amino acids. The amino acids are divided into three classes according to attribute, and each amino acid is encoded by one of the indices 1,2, 3 according to which class it belongs. Table \ref{tab:22} shows that amino acid attributes and corresponding division.

\section{Pre-train Amino Acid Embeddings}

We use the embedding method to represent each amino acid $a\in \mathcal{A}$ as a vector. Each embedding vector is a concatenation of two sub-embedding, i.e. $E(a)=[E_1(a), E_2(a)]$. The first part $E_1$ measures the co-occurrence similarity of the amino acids, obtained by pre-training the Skip-Gram\cite{mikolov2013distributed} model protein sequences. The skip-gram model is trained using negative sampling, where the vocabulary samples are overlapping 3-mer amino acids, and the word vector size is 5. The second part $E_2$ is a one-hot encoding based on the classification defined by the similarity of electrostaticity and hydrophobicity among amino acids, where 20 natural amino acids can be clustered into 7 classes\cite{shen2007predicting}, shown in Table \ref{tab:33}. For the $21^{st}$ amino acid U(Selenocysteine), the $22^{nd}$ amino acid O(Pyrrolysine) and the unknown amino acids X are included in the eighth category. In summary, each amino acid is expressed as $E(a) \in R^{5+(7+1)=13}$.

\begin{table}[htb]
\small
\centering
\begin{tabular}{cccc}
\hline
No. & Dipole scale & Volume scale & Class\\
1 & - & - & A, G, V\\
2 & - & + & I, L, F, P\\
3 & + & + & Y, M, T, S\\
4 & ++ & + & H, N, E, W\\
5 & +++ & + & R, K\\
6 & +'+'+' & + & D, E\\
7 & +'' & + & C\\
\hline
\end{tabular}
\caption{Classification of amino acids. Dipole scale: -, Dipole$<$1.0; +, 1.0$<$Dipole$<$2.0; ++, 2.0$<$Dipole$<$3.0; +++, Dipole$>$3.0; +'+'+', Dipole$>$3.0 with opposite orientation; +'', Cys is separated from class 3 because of its ability to form disulfide bonds. Volume scale: -, Volume$<$50; +, Volume$>$50.}
\label{tab:33}
\end{table}

% In summary, the problem can be formalized as: Given the Protein set $\mathcal{P}$ and PPI set $\mathcal{X}$, where $|\mathcal{P}| = N, |\mathcal{X}| = M$. The PPI network is denoted as $G=(\mathcal{P}, \mathcal{X})=(N, M)$, and assume $G$ is connected(The $G$ used in the original paper are all connected). We use random partition strategy $\mathcal{S}_{r}$ and test size ratio $t$ to divide $G$ into $G_{train}=(\mathcal{P}_{v}, \mathcal{X}_{train})$ and $G_{test}=(\mathcal{P}_{u}, \mathcal{X}_{test})$. We need to prove is under what situation, satifying $|\mathcal{P} - \mathcal{P}_v| \leq \sigma|\mathcal{P}|, \sigma \leq 0.001$. When $\sigma = 0$, $G_{train}$ is connected.

\begin{table}[!htb]
\small
\centering
\begin{tabular}{c|l|r}
\hline
\multicolumn{2}{c|}{Hyper-Parameters} & Values\\
\hline
 & Fixed amino acid length & 2000\\
% \hline
Model & Protein-I Feature & 256\\
Architecture& Protein-G Feature & 50 \\
& Graph layers & 1 \\
\hline
 & learning rate(lr) & 0.001 \\
& lr reduce rate & 0.5 \\
Model & lr reduce patience & 20 \\
Training & l2 weight decay & 5e-4 \\
& batch size & 1024 \\
& epochs & 300 \\
\hline
\end{tabular}
\caption{The hyper-parameter settings for GNN-PPI.}
\label{tab:55}
\end{table}

% \section{Real-World PPI Network}

% We perform empirical studies by comparison of two different time points, 2021/01/25 and 2020/04/11, of the Homo sapiens subset of BioGRID database. Some qualitative results are shown in Figure \ref{fig:picture1}, where green and red nodes denote already and newly discovered proteins, respectively. It can be seen that proteins are not discovered randomly. Instead, the newly discovered proteins exhibit some BFS-like or DFS-like local patterns. This may justify that the proposed partitions are more realistic.

% \begin{figure*}[htb]
% \centering
% \includegraphics[width=15cm]{new_rebuttal3.png}
% \caption{Comparison of two different time points of the Homo sapiens of BioGRID database, where green and red nodes denote already and newly discovered proteins, respectively.}
% \label{fig:picture1}
% \end{figure*}

\section{Proportions of Protein}

Table 2 in main text shows some quantitative analysis of the BFS and DFS partitions in terms of the proportions of BS, ES and NS edges. We also calculate the proportions of nodes in testset-only, both-sets and trainset-only as shown in Table \ref{tab:77}. Compared with conventional random partition, BFS and DFS partitions lead to more ES and NS edges, and less nodes in Both-sets, and thus better evaluate the performance of models for unseen proteins.

\begin{table}[!htb]
\centering
\begin{tabular}{ccccc}
\hline
\multirow{2}*{Datasets} & Partition & \multicolumn{3}{c}{Protein Proportions (\%)}\\
\cline{3-5}
    & Schemes & Trainset-only & Testset-only & Both-sets\\
\hline
\multirow{3}*{SHS27k} & Random  & 41.64 & 6.06 & 52.31\\
                      & BFS     & 60.16 & 9.63 & 30.22\\
                      & DFS     & 63.69 & 5.60 & 30.71\\
\hline
\multirow{3}*{SHS148k} & Random  & 34.79 & 4.20 & 61.01\\
                      & BFS     & 52.54 & 9.62 & 37.84\\
                      & DFS     & 51.51 & 5.72 & 42.77\\
\hline
\multirow{3}*{STRING} & Random  & 13.96 & 1.65 & 84.39\\
                      & BFS     & 31.72 & 5.03 & 63.25\\
                      & DFS     & 26.94 & 4.75 & 68.31\\
\hline
\end{tabular}
\caption{Proportion of proteins in trainset and testset over different datasets and data partition schemes.}
\label{tab:77}
\end{table}

% \begin{table}[!htbp]
% \centering
% \begin{tabular}{|c|c|c|c|}
% \hline
% \multirow{2}*{Multi-label} & \multicolumn{3}{c|}{Dataset(\%)} \\
% \cline{2-4}
%  & SHS27k & SHS148k & STRING \\
% \hline
% Reaction & 41.50 & 40.61 & 51.08 \\
% \hline
% Binding & 52.68 & 52.70 & 67.87 \\
% \hline
% Ptmod & 17.09 & 20.98 & 6.92 \\
% \hline
% Activation & 43.24 & 42.50 & 17.53 \\
% \hline
% Inhibition & 18.45 & 20.20 & 6.58 \\
% \hline
% Catalysis & 45.80 & 44.66 & 48.59 \\
% \hline
% Expression & 9.01 & 7.68 & 2.07 \\
% \hline
% \end{tabular}
% \caption{Proportion of different labels in different datasets.}
% \end{table}

\end{appendices}

\end{document}